\documentclass[journal,twoside,web]{ieeecolor}
\usepackage{tmi}
\usepackage{amsmath,amssymb,amsfonts}
\usepackage{algorithmic}
\usepackage{graphicx}
\usepackage{textcomp}
\usepackage{hyperref}
\usepackage{booktabs}
\usepackage{multirow}
\usepackage{xcolor}

\def\BibTeX{{\rm B\kern-.05em{\sc i\kern-.025em b}\kern-.08em
    T\kern-.1667em\lower.7ex\hbox{E}\kern-.125emX}}
\begin{document}
\title{Self-Supervised Contrastive Embedding Adaptation for Endoscopic Image Matching}

\author{Alberto Rota,  Elena De Momi
\thanks{Alberto Rota and Elena De Momi are with the Department of Electronics, Information and Bioengineering, Politecnico di Milano, 20133 Milan, Italy}
\thanks{Alberto Rota is the corresponding author at \href{mailto:alberto1.rota@polimi.it}{alberto1.rota@polimi.it}}}
\maketitle

\begin{abstract}
Accurate spatial understanding is essential for image-guided surgery, augmented reality integration and context awareness. In minimally invasive procedures, where visual input is the sole intraoperative modality, establishing precise pixel-level correspondences between endoscopic frames is critical for 3D reconstruction, camera tracking, and scene interpretation. However, the surgical domain presents distinct challenges: weak perspective cues, non-Lambertian tissue reflections, and complex, deformable anatomy degrade the performance of conventional computer vision techniques. While Deep Learning models have shown strong performance in natural scenes, their features are not inherently suited for fine-grained matching in surgical images and require targeted adaptation to meet the demands of this domain.

This research presents a novel Deep Learning pipeline for establishing feature correspondences in endoscopic image pairs, alongside a self-supervised optimization framework for model training. The proposed methodology leverages a novel-view synthesis pipeline to generate ground-truth inlier correspondences, subsequently utilized for mining triplets within a contrastive learning paradigm. Through this self-supervised approach, we augment the DINOv2 backbone with an additional Transformer layer, specifically optimized to produce embeddings that facilitate direct matching through cosine similarity thresholding.

Experimental evaluation demonstrates that our pipeline surpasses state-of-the-art methodologies on the SCARED datasets improved matching precision and lower epipolar error compared to the related work. The proposed framework constitutes a valuable contribution toward enabling more accurate high-level computer vision applications in surgical endoscopy.

\end{abstract}


\section{Introduction}
\label{sec:introduction}

Minimally Invasive Surgery (MIS) has revolutionized modern surgical practice by reducing patient trauma, shortening recovery times, and improving cosmetic outcomes. However, these benefits come at the cost of restricted vision, limited tactile feedback, and increased technical complexity. Computer vision systems that can analyze the endoscopic video feed in real-time offer the potential to address these limitations by providing Augmented Reality (AR) overlays, automated tool tracking, and 3D reconstruction of the surgical field \cite{MaierHein2020Surgical, Vercauteren2020CAI4CAI}. 
Computer vision applications in surgery are designed to enhance a surgeon’s situational awareness, offering improved perception of the surgical environment, enhanced spatial awareness, and a deeper understanding of the operative anatomy. These advancements contribute to a more precise, informed, and efficient surgical experience \cite{ward2021computer}: ultimately, the benefits for surgeons translate into benefits for patients, as these applications help reduce operative and recovery time while minimizing intraoperative errors \cite{penza2020enhanced}.

A fundamental requirement for these assistive technologies is the ability to accurately identify and track corresponding points between endoscopic video frames, a task known as feature matching.  Feature matching is the cornerstone of foundational methods like image registration, object recognition, and 3D reconstruction in computer vision applications. In the context of surgery, robust feature matching techniques enable accurate tracking of instruments, visual odometry for camera localization, identification of anatomical structures, and spatial alignment of preoperative imaging with real-time intraoperative views \cite{rieke2018computer}.
Feature matching in endoscopic scenes presents unique challenges not commonly encountered in general computer vision applications, which focus on outdoor and indoor scenery. The surgical environment is characterized by non-rigid tissue deformation, dynamic lighting conditions, specular reflections, smoke, and blood. These factors make traditional image processing methods less effective, requiring more robust approaches such as deep learning-based feature extraction and domain-specific adaptations \cite{kennedy2020computer, cartucho2021visionblender}.

In this research, we present a self-supervised optimization framework for adapting the semantically-rich representations extracted from pretrained vision transformer architectures to generate patch embeddings that facilitate efficient correspondence matching in the endoscopic imaging domain. The primary contributions of this investigation are threefold:

\begin{itemize}
    \item An end-to-end pipeline for feature extraction and matching based on a neural network trained to adapt the extracted semantic embedding map to local matchable features.  
    \item A self-supervised training protocol that exploits novel-view synthesis techniques to compute geometrically consistent pixel correspondences between endoscopic images and their synthetically generated novel viewpoints, thereby circumventing the necessity for manual annotation.
    
    \item A quantitative evaluation framework that compares the performance of our proposed methodology against established state-of-the-art approaches, both those trained on general domain datasets and those specifically optimized for endoscopic imaging applications.
\end{itemize}

\section{Related Works}
Traditional feature-based methods such as Scale-Invariant Feature Transform (SIFT) \cite{Jindal2014sift}, Speeded-Up Robust Features (SURF)\cite{Bay2008surf}, and Oriented FAST and Rotated BRIEF (ORB) \cite{Rublee2011ORB} have demonstrated high accuracy in structured and well-textured environments. However, these approaches perform poorly in MIS, due to the lack of distinctive textures, specular reflections, tissue deformations, and varying illumination conditions \cite{Chu2020endoscopic, Xu2024vascular,Zhang2023dgn}.  To mitigate this, some studies have attempted modifications, such as refining feature descriptors or incorporating topology preservation strategies \cite{Pourshahabi2024fast}. However, these handcrafted features still lack robustness under non-rigid deformations common in MIS.

Another traditional approach is feature matching for visual simultaneous localization and mapping (VSLAM) in surgical navigation. Zhang et al. \cite{Zhang2023vslam} proposed a feature-matching method tailored for VSLAM applications in robotic surgery, demonstrating improved accuracy in mapping deformable tissues. Their approach incorporated motion constraints to refine feature correspondences and was validated on real endoscopic datasets.
\begin{figure*}[h!]
    \centering
    \includegraphics[width=\textwidth]{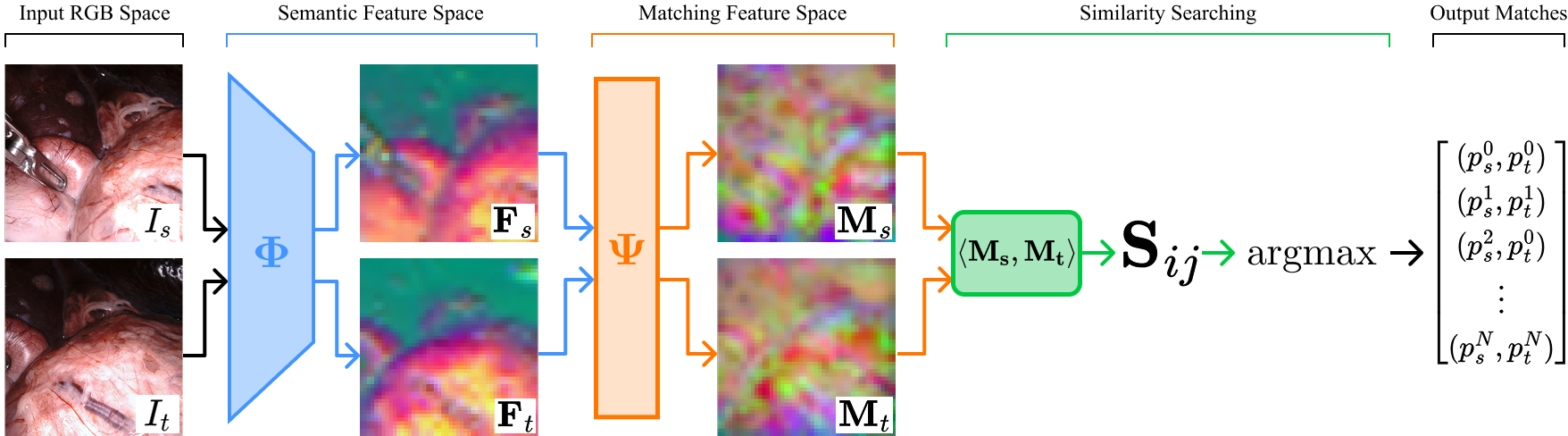}
    \caption{Model architecture overview. The pretrained DINOv2 backbone $\Phi$ extract features in a semantic domain. We introduce a trainable vision transformer layer $\Psi$, optimized to adapt the semantic features to a more discriminable domain, targeted to a similarity searching task. The semantic features $\mathbf{F_s}$ and $\mathbf{F_t}$ and their adaptations $\mathbf{M_s}$ and $\mathbf{M_t}$ are graphically represented as their RGB-encoded principal components. The outputs of the adaptation module $\Psi$ serve as the set of descriptors from which the similarity matrix $\mathbf{S}_{ij}$ is computed and ultimately converted to a set of matching pixel coordinates through $argmax$ operations. 
}
    \label{fig:model}
\end{figure*}
Descriptor learning methods that aim at improving feature robustness and discriminability play a crucial role in computer vision tasks such as image matching and 3D reconstruction. AffNet \cite{affnet} enhances feature matching by employing a neural network to estimate affine transformations, allowing for normalization of local image patches and improving descriptor resilience to viewpoint changes. HyNet \cite{hynet} refines local descriptor learning by incorporating hybrid similarity measures with triplet loss, resulting in more discriminative and robust feature descriptors. SOSNet \cite{sosnet} introduces second-order similarity regularization, leveraging second-order feature statistics to enhance descriptor distinctiveness and reliability. These advancements collectively contribute to more accurate and robust image feature representations, significantly improving performance in challenging visual matching scenarios.

Deep learning has significantly advanced feature-matching techniques also in medical imaging, particularly through Convolutional Neural Networks (CNNs). Zhang et al. \cite{Zhang2023dgn} introduced a Descriptor Generation Network (DGN) for monocular endoscopy, which learns feature representations in a local-to-global manner. The results showed superior feature matching in sparse depth estimation tasks compared to traditional feature-based methods.
Similarly, Li et al. \cite{Li2021monocular} explored monocular tissue reconstruction using a robotic-assisted framework. Their approach employed SIFT and SURF-based multi-level feature matching for reconstructing soft tissues in 3D, addressing challenges in endoscopic depth estimation.

Self-supervised learning (SSL) has emerged as a promising direction for feature learning, with the groundbreaking work from Mishchuk et al. \cite{hardnet8} with HardNet, demonstrating the accuracy gain of contrastive learning with efficient mining.
Farhat et al. \cite{farhat2023self} applied the HardNet paradigm to a self-supervised endoscopic image key-points matching approach that demonstrated state-of-the-art performance in MIS scenarios.
Chu et al. \cite{Chu2020endoscopic} proposed a motion consensus-based feature matching method that integrates affine transformation estimation with bilateral regression. This method improved feature correspondence accuracy in non-rigid endoscopic sequences. Furthermore, Xu et al. \cite{Xu2024vascular} developed a vascular feature detection model leveraging dual-branch structure enhancement, which significantly improved matching robustness in laparoscopic imagery.

Pretrained Vision Transformer (ViT) architectures, particularly DINOv2 \cite{oquab2023dinov2}, have demonstrated strong generalization capabilities in endoscopic image domains \cite{batic2023endoscopy}\cite {sanderson2024selfsupervised}. Integrating a Dense Prediction Transformer (DPT) head \cite{ranftl2021vision} enables the generation of smooth and reasonably accurate depth maps from monocular RGB images, albeit without metric scale nor multi-view consistency. This approach has proven effective in various medical imaging tasks, including depth estimation \cite{veasey2024dinov2}.

While CNNs primarily focus on local feature extraction, ViTs and self-attention mechanism leverage long-range dependencies that can model global context, which is particularly useful in texture-less and repetitive endoscopic environments. Recent studies have shown that transformer-based feature matchers outperform CNN-based methods in complex MIS scenarios, particularly in sparse feature environments and low-texture regions \cite{Zhang2023dgn, Pourshahabi2024fast}. Notably, SuperGlue \cite{superglue} and LoFTR \cite{loftr} have showcased impressive performance in feature matching on indoor and outdoor scene by eliminating the need for keypoint detection and utilizing attention-based mechanisms for feature correspondence. Lu et al. \cite{lu2024s2p} utilized a vision transform architecture for matching images from capsule endoscopy videos.

Despite these advancements, feature matching in MIS remains an open challenge. Most existing methods still face issues related to real-time performance, with transformer-based methods often requireing large-scale surgical datasets for effective training \cite{Xu2024vascular}. 

\section{Methodology}
\label{sec:methods}
\subsection{Model Architecture}
The main contribution of this paper is a descriptor extraction model \( \Gamma \), which takes as input a pair of endoscopic RGB images, denoted as the \textit{source} and \textit{target} images \( I_s \) and \( I_t \), and outputs a set of \( N \) descriptor vectors each of length \( E \):
\begin{equation}
\mathbf{M}_s = \Gamma(I_s; \theta), \quad \mathbf{M}_t = \Gamma(I_t; \theta)
\end{equation}

where \( \mathbf{M}_s, \mathbf{M}_t \in \mathbb{R}^{N \times E} \), and \( \theta \) denotes the set of learnable model parameters. 

A set of \( N \) pixel correspondences \( \{(p_s^i, p_t^i)\}_{i=1}^N \), where each pair consists of matching coordinates in the source and target images, is obtained by comparing descriptor similarity, specifically finding the maximums of a similarity function \( S(\cdot, \cdot) \):

\begin{equation}
\{(p_s^i, p_t^i)\}_{i=1}^N  = \arg\max \{ S(\mathbf{M}_s, \mathbf{M}_t) \}
\end{equation}

\label{sec:model}
We adopt the pre-trained DINOv2 model \cite{oquab2023dinov2} as our baseline feature extractor for obtaining descriptors. While DINOv2 offers strong semantic representations, its direct application to feature matching in surgical imagery is limited. The model is optimized to cluster semantically similar regions in feature space, often ignoring spatial or geometric distinctions. This behavior impairs its suitability for pixel-level correspondence, where local distinctiveness and geometric consistency are essential. Therefore, adapting DINOv2 is necessary to align its feature space with the spatial precision required in endoscopic matching tasks.

For the proposed adaptation, given a pair of input images $I_s, I_t$ the DINOv2 backbone $\Phi$ independently processes each image:
\begin{equation}
\mathbf{F}_s = \Phi(I_s), \quad \mathbf{F}_t = \Phi(I_t)    
\end{equation}
where $\mathbf{F}_s, \mathbf{F}_t \in \mathbb{R}^{E \times \frac{H}{P} \times \frac{W}{P}}$ are the resulting semantic feature maps. Here, $H$ and $W$ are the image height and width, $E$ represents the embedding dimension (768 for DINOv2-base variant), and $P$ denotes the patch size (14 for DINOv2). 
Feature maps therefore have a resolution of $P\times P$ pixels each, corresponding to a total of $N = \frac{H \cdot W}{P^2}$ descriptors per image.

The proposed architecture enhances the baseline model by incorporating a dedicated transformer module, denoted as \( \Psi \), specifically optimized for pixel-level correspondence estimation (see Figure~\ref{fig:model}). 
\( \Psi \) adapts $\mathbf{F}_s$ and $\mathbf{F}_t$ to produce a set of discriminative matching descriptors \( \mathbf{M}_s, \mathbf{M}_t \in \mathbb{R}^{N \times E} \), as follows:
\begin{equation}
\mathbf{M}_s = \Psi(\mathbf{F}_s), \quad \mathbf{M}_t = \Psi(\mathbf{F}_t),
\end{equation}
The purpose of this adaptation layer is to facilitate the transformation of DINOv2's semantically-enriched embeddings into a feature space characterized by heightened local distinctiveness and discriminative capacity. 
The adaptation module \( \Psi \) is a vision transformer \cite{dosovitskiy2020image} layer. 

The resulting descriptors \( \mathbf{M}_s \) and \( \mathbf{M}_t \) are subsequently used as input to a similarity function for establishing dense correspondences. Specifically, we implement a similarity searching module that establishes correspondences between patches in the source and target images. We compute a similarity matrix $\mathbf{S} \in \mathbb{R}^{N \times N}$ between all pairs of descriptors:
\begin{equation}
\mathbf{S}_{ij} = \frac{\mathbf{M}_s^i \cdot \mathbf{M}_t^j}{||\mathbf{M}_s^i|| \cdot ||\mathbf{M}_t^j||}
\end{equation}

This similarity matrix encodes the likelihood of correspondence between all pairs of descriptors across the image pair. Formally, a correspondence $(i,j)$ is retained if:

\begin{equation}
\arg\max_k \mathbf{S}_{ik} = j \quad \text{and} \quad \arg\max_l \mathbf{S}_{lj} = i
\end{equation}
The confidence score $s_{ij}$ for each correspondence is derived from the similarity value $\mathbf{S}_{ij}$. We retain only matches with an empirical similarity value $s_{ij} >0.95$.
For each patch correspondence $(\mathcal{P}_s^i, \mathcal{P}_t^i)$, subpixel refinement is accomplished via phase correlation in the frequency domain. Corresponding patches undergo Fourier-based correlation:

\begin{equation}
C(u,v) = \mathcal{F}^{-1}\left\{\frac{\mathcal{F}\{\mathcal{P}_s\} \cdot \mathcal{F}\{\mathcal{P}_t\}^*}{|\mathcal{F}\{\mathcal{P}_s\} \cdot \mathcal{F}\{\mathcal{P}_t\}^*|}\right\}
\end{equation}

where $C$ is a spatial correlation map, $\mathcal{F}$ and $\mathcal{F}^{-1}$ represent the Fourier transform and its inverse and $*$ denotes complex conjugation. The displacement vector 
\begin{equation}
    (\Delta u, \Delta v) = \arg\max_{u,v} C(u,v)
\end{equation}
is identified from the correlation peak, enabling correspondence refinement: $p_s^i \leftarrow p_s^i + (\Delta u, \Delta v)$. This affords subpixel precision essential for geometric estimation procedures.

\begin{figure*}[h!]
    \centering
    \includegraphics[width=\textwidth]{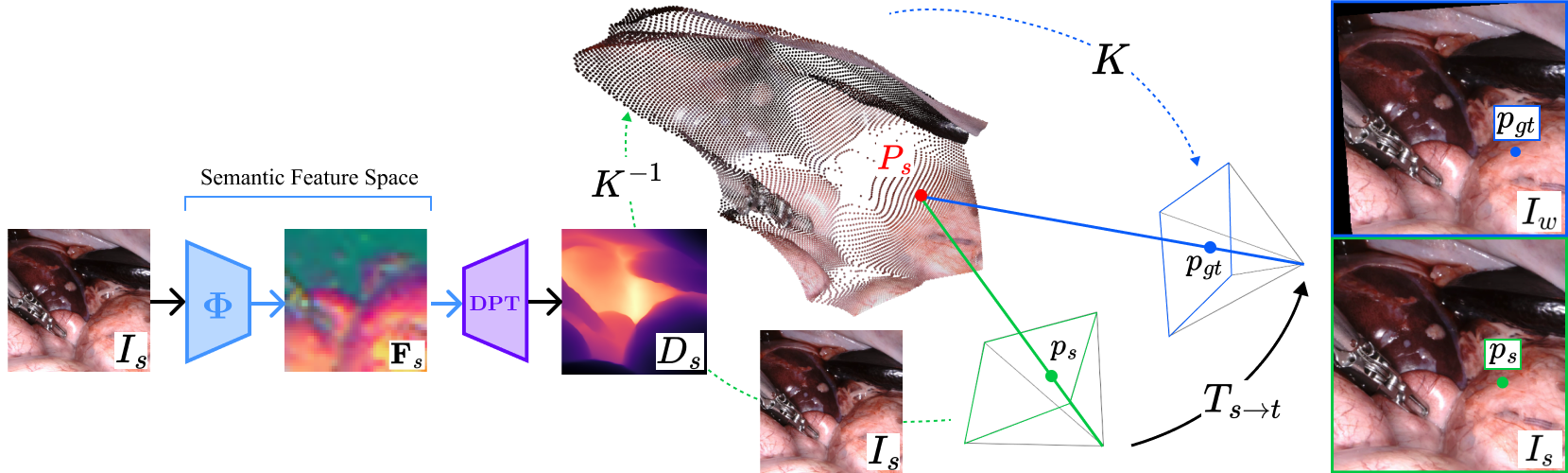}
    \caption{Pipeline for constructing a novel view $I_w$ from a source image $I_s$ obtaining a set of pseudo-ground-truth pixel correspondences $\{(p_s^i, p_{gt}^1)\}_{i=1}^N$. The source image is back-projected through the inverse intrinsic camera parameters into the RGB-3D space (green dashed arrow). The depthmap $D_s$ required for this back-projection is obtained decoding the semantic features $\mathbb{F}_s$ with DPT. The resulting point cloud is transformed to the reference frame of the target image with the transformation matrix $T_{s \rightarrow t}$ and then projected to image space via the intrinsics, synthesizing a novel \textit{warped} view $I_w$ of the source image (blue dashed arrow). This procedure enables systematic mapping of any source image pixel $p_s$ to its corresponding position in the novel view $p_{gt}$ via projection through the intrinsics (blue dashed arrow).}
    \label{fig:projections}
\end{figure*}

\subsection{Self-supervision from Novel-View Synthesis}

The adaptation module $\Psi$ is a learned function optimized with a self-supervised approach that leverages novel view synthesis to generate pseudo-ground-truth pixel correspondences between image pairs. This methodology eliminates the necessity for annotated ground-truth correspondences.
The core insight of this approach lies in the exploitation of geometric constraints derived from multi-view geometry to synthesize correspondence data. Figure \ref{fig:projections} illustrates the whole pipeline for synthesizing such correspondences, which is described below.

Given a source image \( I_s \) acquired by a camera at a generic pose \( T_s \), a known relative pose transformation \( T_{s \rightarrow t} \in SE(3) \), and the camera intrinsic matrix \( K \in \mathbb{R}^{3 \times 3} \), it is possible to predict the pixel coordinates of any feature in the image plane of a target camera pose \( T_t \) by back-projecting to 3D space and projecting to image space, provided that depth information is available.

Let $p_s = (u_s, v_s, 1)^\top$ denote a pixel in homogeneous coordinates in the source image. With a depth map $D_s$ for the source image, this pixel can be back-projected to 3D space:
\begin{equation}
P_s = D_s(p_s) \cdot K^{-1}p_s
\end{equation}
where $P_s \in \mathbb{R}^3$ represents the corresponding 3D point in the source camera's coordinate frame. To project this point into the target view, the relative rigid camera transformation in the form of the homogenous matrix $T_{s \rightarrow t}$ is applied:

\begin{equation}
P_t = T_{s \rightarrow t} \cdot \begin{pmatrix} P_s \\ 1 \end{pmatrix}
\end{equation}

Finally, this 3D point is projected onto the target image plane, intercepting it at the pseudo-ground-truth position $p_{gt}$:

\begin{equation}
p_{gt} = \pi(K \cdot P_t)
\end{equation}

where $\pi$ is the perspective projection function that converts a 3D point in camera coordinates to pixel coordinates:

\begin{equation}
\pi\begin{pmatrix} X \\ Y \\ Z \end{pmatrix} = \begin{pmatrix} X/Z \\ Y/Z \\ 1 \end{pmatrix}
\end{equation}

\begin{figure*}[h!]
    \centering
    \includegraphics[width=\textwidth]{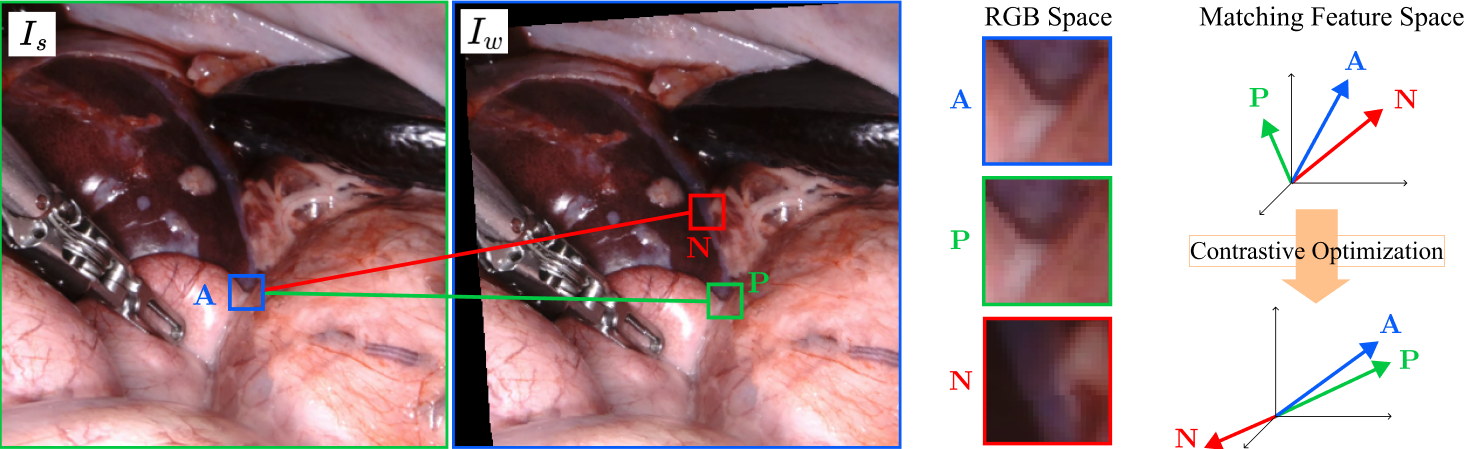}
    \caption{Triplet mining pipeline with an example image pair and one triplet. In feature space (visualized using 3 principal components), the anchor embedding initially shows similarity to the negative sample. Contrastive optimization subsequently aligns the anchor closer to the positive sample while increasing its distance from the negative sample, enhancing the discriminative capacity of the learned representations.}
    \label{fig:triplets}
\end{figure*}

This process establishes a bi-univocal correspondence between pixel $p_s$ in the source image and $p_{gt}$, the corresponding position on the target image plane given the depthmap and the homogeneous camera transformation. By assigning to each pixel in $p_{gt}$ the RGB color values from the corresponding $p_s$, with a z-buffering strategy to handle overlapping projections, this approach synthesizes a novel \textit{warped} view $I_w$ of the source RGB image, and procedurally obtains the pseudo-ground-truth $p_{gt}$ position for any $p_s$ pixel in the source image. 

In practice, a metric depth map is rarely available and must typically be estimated using monocular depth estimation models, which produce depth maps that are \textit{up-to-scale}—that is, not in real-world metric units.
This limitation does not invalidate the approach: for the purpose of computing pseudo-ground-truth correspondences, a high-quality, smooth and depth-preserving depth map with an unknown scale is arguably sufficient. Moreover, assigning a random (but sensible) scaling factor to these depth maps can enhance the diversity of the supervision signal, thereby improving the model’s generalization capability.

This considered, we rely on the Dense Prediction Transformer (DPT)~\cite{ranftl2021vision} to decode the semantic feature map \( \mathbf{F}_s \) into a depth map \( D_s \in \mathbb{R}^{1 \times H \times W} \). 
The DPT architecture, pre-trained for monocular depth estimation, is designed to anticipate feature distributions consistent with those produced by the frozen DINOv2 backbone. Therefore, altering the foundational DINOv2 layers would potentially degrade depth estimation performance, thereby weakening the supervision signal derived from view synthesis.
To preserve this compatibility, the proposed vision transformer layer \( \Psi \) leaves the pre-trained backbone weights unchanged, while introducing a trainable transformation that maps general-purpose visual representations into an embedding space optimized for pixel-level correspondence.

The depth map \( D_s \)—used exclusively during training—is derived from the frozen features \( \mathbf{F}_s \) and is not influenced by gradient updates. Consequently, the model backbone operates in dual capacities: it produces semantically-rich representations \( \mathbf{F}_s, \mathbf{F}_t \) leveraged for depth estimation at training time, and correspondence-optimized embeddings \( \mathbf{M}_s, \mathbf{M}_t \) generated via \( \Psi \), which are tailored for the matching task at inference time. Notably, the shapes and resolutions of \( \mathbf{M}_s \) and \( \mathbf{M}_t \) are identical to those of \( \mathbf{F}_s \) and \( \mathbf{F}_t \).

Similar considerations apply to the camera motion matrix \( T_{s \rightarrow t} \): although several datasets~\cite{allan2021stereo, hayoz2023pose} provide video sequences with synchronized camera motion trajectories, it is not necessary to use the exact metric transformations from the dataset. Instead, the rotational and translational components of \( T_{s \rightarrow t} \) can be randomly sampled within a realistic range to exponentially diversify the supervision signal and improve the robustness of the learned correspondence model.

\subsection{Contrastive Learning}
The additional vision transformer layer $\Psi$ introduced above is trained utilizing contrastive learning methodologies, which facilitate the projection of corresponding RGB features into proximal locations within the embedding space. The descriptor extraction architecture generates feature maps with coarse spatial resolution of dimensions $\frac{H}{P} \times \frac{W}{P}$, whereas the supervisory signal is derived from pixel-level correspondences between the source and warped images. Consequently, for any given pixel-level correspondence $\{(p_s^i, p_t^i)\}$, we define the associated patch-level correspondence as the pair $\{(\mathcal{P}_s^i, \mathcal{P}_t^i)\}$, where $\mathcal{P}_s^i$ and $\mathcal{P}_t^i$ represent the $P \times P$ RGB patches that $p_s^i$ and $p_t^i$ respectively belong to. 

The contrastive learning framework is implemented through the gradient-based minimization of the triplet loss function. The triplet loss criterion facilitates the optimization of a feature space wherein corresponding image regions are situated in proximity, while non-corresponding regions maintain sufficient separation:

\begin{equation}
\mathcal{L}_{\text{triplet}} = \max(0, d(\mathbf{A}, \mathbf{P}) - d(\mathbf{A}, \mathbf{N}) + m)
\end{equation}

where $d(\cdot, \cdot)$ represents the cosine distance between embeddings, defined as:

\begin{equation}
d(\mathbf{u}, \mathbf{v}) = \sqrt{2 - 2 \cdot \frac{\mathbf{u}^T \mathbf{v}}{||\mathbf{u}|| \cdot ||\mathbf{v}||}}
\end{equation}

and $m$ is a positive scalar margin set to 1 as in the original HardNet implementation \cite{hardnet8}.
A triplet is defined as a set of three embedding descriptors, each of shape $\mathbb{R}^E$: an anchor $\mathbf{A}$, a positive $\mathbf{P}$, and a negative $\mathbf{N}$. These constituents are selected according to the following protocol:

\begin{enumerate}
   \item \textbf{Anchor Selection}: An anchor embedding $\mathbf{A}$ extracted from the generic patch $\mathcal{P}_s^i$ in the source image
   \item \textbf{Positive Selection}: The positive embedding $\mathbf{P}$ represents the patch $\mathcal{P}_t^i$ in the warped image that contains $p_{gt}$. This correspondence is established by synthesizing the novel view.
   \item \textbf{Negative Selection}: The negative embedding $\mathbf{N}$ is determined via a principled hard-negative mining strategy that maximizes the discriminative signal. We formulate this selection mathematically as:
    \begin{equation}
    \mathbf{N} = \arg\min_i \{ d(\mathbf{A}, \mathbf{M}_t^i), d(\mathbf{P}, \mathbf{M}_t^i) \} 
    \end{equation}
    where $\mathbf{M}_t^i \in \{\mathbf{M}_t\}$.
    The negative embedding $\mathbf{N}$ is hence chosen as the most difficult to discriminate from the positive in feature space, and is the embedding in $\mathbf{M}_t$ that is either closest to the anchor or the positive.
\end{enumerate}

Figure \ref{fig:triplets} provides a visualization of an example triplet. For a batch of source-warped image pairs gathered from the novel view synthesis pipeline, any given number of triplets can be mined up to memory constraints. However, we introduce a cycle-consistency constraint on Anchor-Positive matches.

Gradients computed from $\mathcal{L}_{\text{triplet}}$ are back-propagated only to the adaptation transformer layer $\Psi$ attached to the pretrained sequence.

\subsection{Dataset}
\label{sec:dataset}

The training and empirical validation of our methodology utilizes splits of the Stereo Correspondence And Reconstruction of Endoscopic Data (SCARED) \cite{allan2021stereo}, which comprises 9 stereoscopic video sequences captured during ex-vivo porcine procedures. Each sequence contains approximately 300-900 stereo frames at 1280$\times$1024 pixel resolution, with camera calibration parameters and ground-truth camera poses derived from robotic kinematics. The dataset exhibits characteristic surgical challenges including specular reflections and instrument occlusions. We partition the dataset into 7 sequences for training and validation, and 2 for testing. We retain only the left image from the stereo sequence.

\paragraph{Training Phase}
At training time, a single \textit{source} image is sampled from a video sequence. A \textit{warped} synthetic view is then generated by applying a randomly sampled rigid camera transformation matrix, drawn from the empirical distribution of relative transformations $T_{s \rightarrow t}$ observed in the dataset. Although the SCARED dataset provides ground-truth camera poses, we do not constrain the training process to use the true pose corresponding to the target frame. This approach allows us to generate a significantly larger number of diverse image pairs. Note that the warped synthetic view may contain artifacts such as unprojected regions or occlusions. To further improve robustness and encourage feature invariance, we also apply random brightness and saturation shifts to the warped view during training.

\paragraph{Inference Phase}
At test time, we sample a sequential pair of frames: a \textit{source} and a \textit{target} image. No synthetic view is generated during inference; the method operates directly on the real image pair.

\begin{figure*}
    \centering
    \includegraphics[width=\linewidth]{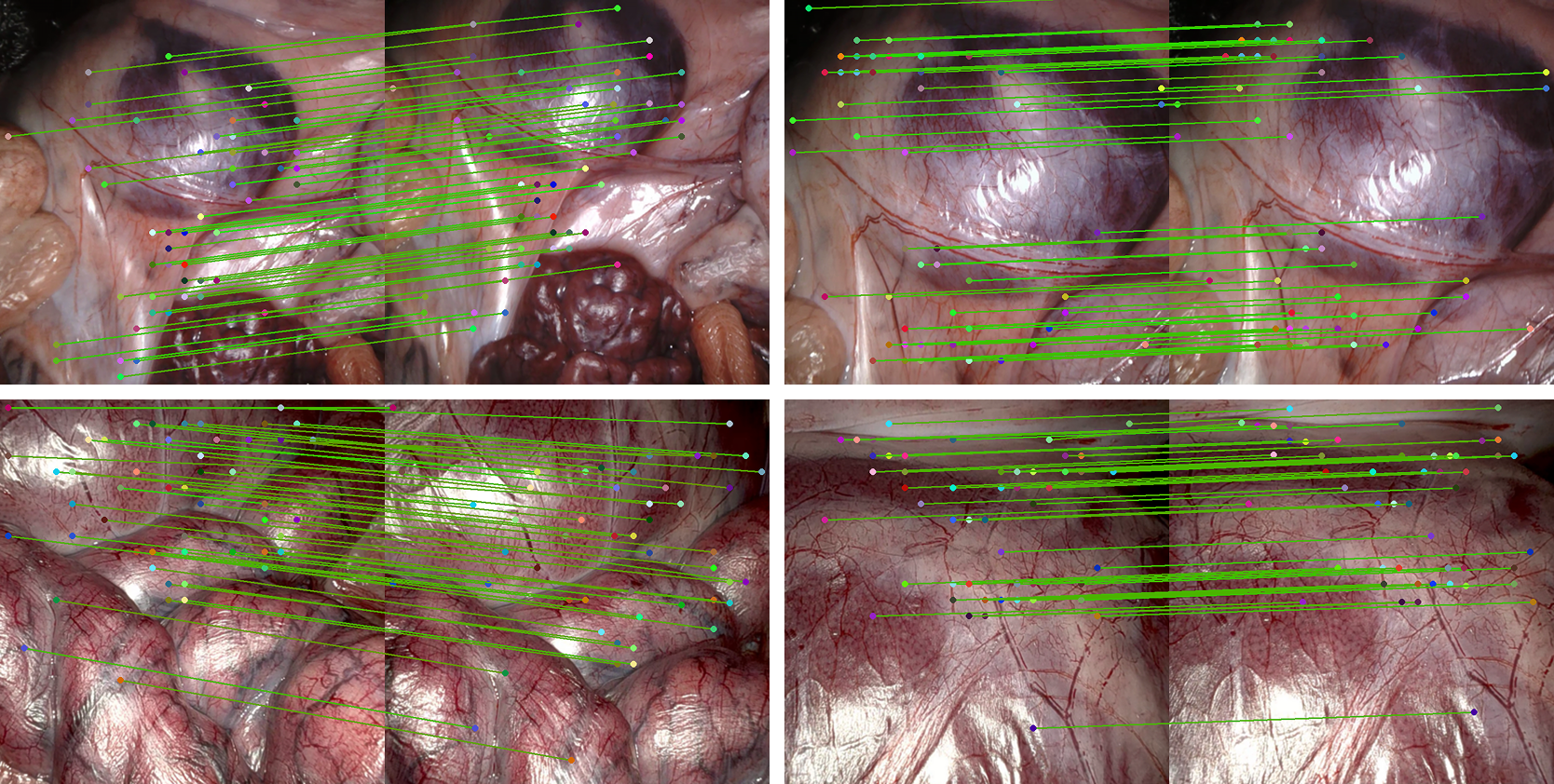}
    \caption{Qualitative visualization of pixel correspondences generated by our model. Four representative examples from the 2 sequences in the SCARED test split are displayed. For each sequence, we present the two image pairs with the highest inlier count, highlighting the top 50 highest-scoring matches between pairs to demonstrate the model's accuracy across varying surgical scenes.}
    \label{fig:examples}
\end{figure*}

\subsection{Evaluation Metrics}
\label{sec:metrics}
We compare the matching performance of our model to the related work on the test split of the SCARED dataset. SCARED does not provide ground-truth matches between adjacent frames, but it provides the $T_{s\rightarrow t}$ rigid camera pose transformation matrix between frame pairs and the intrinsic camera parameter $K$ for each frame. With this information we obtain the ground-truth fundamental matrix $F_{gt}$ as

$$F_{gt} = K^{-\top} [t]_{\times} R K^{-1}$$

where $R \in \mathbb{R}^{3 \times 3}$ and $t \in \mathbb{R}^3$ are the rotational and translational components of $T_{s\rightarrow t}$, respectively. The operator $[t]_{\times}$ generates the skew-symmetric matrix corresponding to the cross-product with $t$.

The fundamental matrix encodes the epipolar geometry of the scene and is uniquely determined from the rigid camera motion and intrinsic parameters. We leverage $F_{gt}$ by evaluating the model performance on the following quantitative metrics:
\paragraph{Epipolar Error}
Quantifies the mean symmetric epipolar distance between corresponding points and their respective ground-truth epipolar lines:

\begin{equation}
\begin{split}
E_e = \frac{1}{|\mathcal{C}|} \sum_{(p_s^i, p_t^i) \in \mathcal{C}} 
\left( 
\frac{|(p_t^i)^\top F_{gt} p_s^i|}
{\sqrt{(F_{gt}p_s^i)_1^2 + (F_{gt}p_s^i)_2^2}} 
\right. \\
\left. + \frac{|(p_s^i)^\top F_{gt}^\top p_t^i|}
{\sqrt{(F_{gt}^\top p_t^i)_1^2 + (F_{gt}^\top p_t^i)_2^2}} \right)
\label{eq:epi}
\end{split}
\end{equation}

\paragraph{Precision} 
Quantifies the percentage of True Positive matches among all matches. A match is considered a True Positive (TP) if the distance to its corresponding ground-truth epipolar line is less than 1px (Eq. \ref{eq:epi} is used), \textit{i.e.} it lies on the epipolar line, and it is considered a False Positive (FP) otherwise
\begin{equation}
Pr = \frac{TP}{TP + FP}
\end{equation}

\paragraph{Fundamental Matrix Error}
Measures the disparity between the estimated fundamental matrix $F$ and the ground-truth fundamental matrix $F_{\text{gt}}$ using the normalized Frobenius norm:

\begin{equation}
F_e = \frac{\|F - F_{\text{gt}}\|_F}{\|F\|_F \cdot \|F_{\text{gt}}\|_F}
\end{equation}


\paragraph{Inlier Percentage}
Represents the proportion of correspondences that align with the epipolar geometry defined by the fundamental matrix $F_{gt}$. We rely on RANSAC \cite{fischler1981random} to robustly select the inliers $M_{RANSAC}$ from the whole set of matches $M_{all}$

\begin{equation}
I_{\%} = \frac{M_{RANSAC}}{M_{all}}
\end{equation}

\paragraph{Computational Time}
The amount of time in milliseconds for extracting matching pixels from one pair of images.

\subsection{Optimization}
\label{sec:opt}
We conduct our experiments on an NVIDIA V100 GPU via the Adam optimizer for a maximum 100 epochs with discriminative learning rates. During an initial bootstrap phase (epochs 1-3), we apply a learning rate of $1 \times 10^{-4}$ to all model parameters, transitioning subsequently to a refined optimization phase wherein the learning rate is reduced to $5 \times 10^{-5}$ for the remainder of the training regimen.

\section{Experiments and Results}
\subsection{Matching Accuracy}
\begin{table*}[ht]
\centering
\caption{Comparative analysis of the methodology proposed herein juxtaposed with extant approaches across the evaluation metrics delineated in Section \ref{sec:metrics}. The symbol [$\downarrow$] denotes metrics for which lower values indicate superior performance, whereas [$\uparrow$] signifies metrics where higher values are preferable. Values representing optimal performance are emphasized in bold typeface. Statistical significance from a T-Test comparing each method to OURS: $ns$: not significant, *: $p<0.05$, **: $p<0.01$, ***: $p<0.001$.}
\label{tab:metrics}
\begin{tabular*}{\linewidth}{@{\extracolsep{\fill}}lccccc}
\toprule
\multirow{1}{*}{\textbf{Method}} & \textbf{Epipolar Err. [px]} [$\downarrow$] & \textbf{Precision [\%]} [$\uparrow$] & \textbf{Fundamental Err. [\%]} [$\downarrow$] & \textbf{Inliers [\%]} [$\uparrow$] & \textbf{Time [ms]} [$\downarrow$] \\
\midrule
Farhat \textit{et al.} \cite{farhat2023self} & 69.36 $\pm$ 1.75 (***) & 1.36 $\pm$ 0.23 (***) & 44.23 $\pm$ 12.91 (***) & 34.00 $\pm$ 0.26 (***) & 0.63 $\pm$ 0.02 (***) \\
KeyNet \cite{keynet}                & 10.65 $\pm$ 7.08 (**)   & 10.04 $\pm$ 8.24 (***)  & 23.22 $\pm$ 8.08 (ns)  & 97.22 $\pm$ 1.75 (***) & 0.44 $\pm$ 0.03 (***) \\
SuperGlue \cite{superglue}          & 10.89 $\pm$ 7.24 (**)   & 10.40 $\pm$ 8.84 (***)  & 21.32 $\pm$ 7.40 (***) & 99.21 $\pm$ 0.62 (***) & 0.50 $\pm$ 0.00 (***) \\
LoFTR-indoor \cite{loftr}           & 10.78 $\pm$ 7.49 (**)   & 10.39 $\pm$ 8.46 (***)  & 21.19 $\pm$ 5.69 (***) & 99.31 $\pm$ 1.09 (***) & 0.54 $\pm$ 0.01 (***) \\
LoFTR-outdoor \cite{loftr}          & 10.85 $\pm$ 7.57 (**)   & 10.46 $\pm$ 9.31 (***)  & 23.89 $\pm$ 6.13 (ns)  & \textbf{99.76 $\pm$ 0.25} (***) & 0.55 $\pm$ 0.01 (***) \\
SIFT+AffNet \cite{Jindal2014sift}\cite{affnet} & 10.42 $\pm$ 6.69 (*)    & 10.08 $\pm$ 8.70 (***)  & 21.72 $\pm$ 6.88 (***) & 98.14 $\pm$ 1.07 (***) & 0.33 $\pm$ 0.06 (***) \\
SIFT+Hardnet8 \cite{Jindal2014sift}\cite{hardnet8} & 10.45 $\pm$ 6.77 (*)    & 10.06 $\pm$ 8.68 (***)  & 21.99 $\pm$ 7.12 (**)  & 98.24 $\pm$ 1.13 (***) & 0.28 $\pm$ 0.05 (***) \\
SIFT+HyNet \cite{Jindal2014sift}\cite{hynet}      & 10.69 $\pm$ 6.91 (**)   & 9.79 $\pm$ 8.17 (***)   & \textbf{20.62 $\pm$ 7.12} (***) & 96.34 $\pm$ 1.96 (***) & 0.29 $\pm$ 0.06 (***) \\
SIFT+SOSNet \cite{Jindal2014sift}\cite{sosnet}    & 10.64 $\pm$ 6.84 (**)   & 9.80 $\pm$ 8.27 (***)   & 22.89 $\pm$ 6.38 (*)   & 96.51 $\pm$ 1.76 (***) & \textbf{0.27 $\pm$ 0.05} (***) \\
\midrule
\textbf{OURS}                       & \textbf{8.63 $\pm$ 4.14} & \textbf{14.22 $\pm$ 4.61} & 24.54 $\pm$ 5.41       & 75.63 $\pm$ 2.09       & 0.40 $\pm$ 0.00       \\
\bottomrule
\end{tabular*}
\end{table*}

We compare the performance of our self-supervised model with the related work: we involve both deep learning based methods (SuperGlue \cite{superglue}, LoFTR \cite{loftr}, and the method from Farhat \textit{et al.} \cite{farhat2023self}) and standard detector+descriptor methods (SIFT \cite{Jindal2014sift} with AffNet \cite{affnet}, HardNert8 \cite{hardnet8}, HyNet \cite{hynet}, SOSNet \cite{sosnet} and KeyNet \cite{keynet}. The method from Farhat \textit{et al.} is the only one that was trained on endoscopic data. 
The quantitative evaluation metrics are reported in Table \ref{tab:metrics} along with the computational time required for matching a pair of $512 \times 640$ pixel images.
We sample the image pairs in the SCARED test split with a temporal distance of 16 frames between source and target frames, which provides a reasonable balance between local deformation and viewpoint changes.

The results presented in Table \ref{tab:metrics} demonstrate that our approach outperforms existing methods on key metrics. Most notably, our method achieves the lowest mean epipolar error of $8.63 \pm 4.14$ pixels, representing a substantial improvement of approximately 17\% over the next best approach (SIFT+AffNet at $10.42 \pm 6.69$ pixels). Furthermore, our method exhibits the highest precision ($14.22\%$), surpassing all other techniques by a significant margin.

Our approach does not achieve the best performance across all metrics, with SIFT+HyNet showing marginally lower fundamental matrix error and LoFTR-outdoor yielding higher inlier percentage. The computational efficiency of our approach is competitive, requiring only 0.40ms per image pair (the lowest among deep learning methods), which is suitable for real-time applications in surgical navigation. 

Figure \ref{fig:examples} shows four example test images matched with the model trained as described in this research work, for a qualitative visual inspection.

\subsection{Robustness Study}
We evaluate the robustness of our method against false positive matches by employing a challenging cross-sequence matching protocol. In this evaluation, we deliberately sample pairs of images from the SCARED test split with the constraint that source and target images must belong to different video sequences. Under these conditions, the absence of common anatomical structures between views should theoretically result in no valid matches; consequently, we quantify robustness by the inlier ratio, where a more robust model will detect fewer inliers.

In the absence of ground-truth fundamental matrices, we again rely on RANSAC  to estimate an epipolar geometry and filter the outliers that do not fit it.
The results of this comparative analysis are presented in Table \ref{tab:robustness}, which demonstrates that our method produces fewer total spurious matches (1738) in this challenging scenario. For the method proposed in this paper, most detected correspondences are correctly identified as outliers by RANSAC, with only 160 correspondences (9.20\% of the total count) being spuriously classified as inliers.
In this test, the next best performing method from Farhat \textit{et al.} \cite{farhat2023self} detected 200 inliers and 2035 matches. The evaluation reveals that current state-of-the-art methods such as SuperGlue \cite{superglue} and LoFTR \cite{loftr} confidently yield more false positives. 

\begin{table}[h]
\centering
\setlength{\tabcolsep}{0.4em}
\caption{Comparison total match counts and of inlier counts on non-corresponding images across different methods. In both cases, lower is better [$\downarrow$]}
\label{tab:robustness}
\begin{tabular*}{\linewidth}{@{\extracolsep{\fill}}lcc@{}}
\toprule
\textbf{Method} & \textbf{N$^o$ Matches [adim.] [$\downarrow$]} & \textbf{N$^o$ Inliers [adim.] [$\downarrow$]} \\
\midrule
Farhat \cite{farhat2023self} & 2035 & 200 \\
KeyNet \cite{keynet} & 3597 & 878 \\
SuperGlue \cite{superglue} & 6554 & 494 \\
LoFTR indoor \cite{loftr} & 8645 & 3979 \\
LoFTR outdoor \cite{loftr} & 8354 & 1439 \\
SIFT+AffNet \cite{Jindal2014sift}\cite{affnet} & 6340 & 2788 \\
SIFT+Hardnet8 \cite{Jindal2014sift}\cite{hardnet8} & 7129 & 3507 \\
SIFT+HyNet \cite{Jindal2014sift}\cite{hynet} & 7253 & 3517 \\
SIFT+SOSNet \cite{Jindal2014sift}\cite{sosnet} & 6987 & 3269 \\
\midrule
\textbf{OURS} & \textbf{1738} & \textbf{160} \\
\bottomrule
\end{tabular*}
\end{table}

\subsection{Ablation Studies}
\label{sec:ablations}
To elucidate the efficacy of individual components within our proposed methodology, we conducted a series of ablation studies. 

In our baseline configuration, we extract matches utilizing the semantic descriptors $\mathbf{F}_s$ and $\mathbf{F}_t$ derived from the pretrained DINOv2 backbone $\Phi$ without any adaptation layers or pixel-level refinement procedures.
Subsequently, we systematically investigated the contributions of several key enhancements: (1) the integration of the adaptation layer $\Psi$, (2) the implementation of pixel-level refinement techniques $\arg\max_{u,v} C(u,v)$, and (3) the resampling of embedding feature maps to twice their original resolution. 
This latter enhancement was implemented via bilinear interpolation of the DINOv2 feature maps, effectively reducing the patch size from $P \times P$ to $\frac{P}{2} \times \frac{P}{2}$. The underlying rationale is that increasing the spatial resolution of the feature maps could enable each descriptor to represent a more localized image region, thereby improving the spatial precision of the resulting correspondences.

The quantitative outcomes of these ablation studies are presented in Table \ref{tab:component_ablation}.
\begin{table}[h!]
\centering
\caption{Component ablation study results demonstrating the contribution of individual elements in our methodology. }
\label{tab:component_ablation}
\begin{tabular*}{\linewidth}{@{\extracolsep{\fill}}lcc@{}}
\toprule
\textbf{Configuration} & \textbf{Epipolar Err. [px]} [$\downarrow$] & \textbf{Precision [\%]} [$\uparrow$] \\
\midrule
Baseline (DINOv2 only) & 32.74 $\pm$ 12.32 & 1.81 $\pm$ 1.95 \\
+ Adaptation layer & 11.23 $\pm$ 5.87 & 11.45 $\pm$ 4.21 \\
+ Pixel refinement & \textbf{8.63 $\pm$ 4.14} & \textbf{14.22 $\pm$ 4.61} \\
+ Feature resampling ($\frac{P}{2}$) & 9.86 $\pm$ 4.92 & 12.78 $\pm$ 4.43 \\
\bottomrule
\end{tabular*}
\end{table}

The results of the ablation show that upsampling feature maps does not improve the key evaluation metrics.

Furthermore, we examined the differential impact of utilizing either randomly sampled transformations $T_{s\rightarrow t}$ during training or employing the ground-truth camera poses extracted directly from the dataset, and we report a comparison of epipolar error and precision in Table \ref{tab:training_strategy_ablation}. 
\begin{table}[h!]
\centering
\caption{Impact of different training strategies on network performance.}
\label{tab:training_strategy_ablation}
\begin{tabular*}{\linewidth}{@{\extracolsep{\fill}}lcc@{}}
\toprule
\textbf{Training Strategy} & \textbf{Epipolar Err. [px]} [$\downarrow$] & \textbf{Precision [\%]} [$\uparrow$] \\
\midrule
Ground-truth poses & 9.37 $\pm$ 4.96 & 13.11 $\pm$ 4.85 \\
Random pose sampling & \textbf{8.63 $\pm$ 4.14} & \textbf{14.22 $\pm$ 4.61} \\
\bottomrule
\end{tabular*}
\end{table}
These results demonstrate that randomly sampling rigid pose transformations during training to construct synthetic ground-truth for supervision is not only justified but beneficial. 
\section{Discussion}
The proposed method demonstrates competitive performance when compared to a broad range of state-of-the-art techniques, including both learned (e.g., LoFTR~\cite{loftr}, SuperGlue~\cite{superglue}) and handcrafted (SIFT~\cite{Jindal2014sift}) matching pipelines. As detailed in Table~\ref{tab:metrics}, our approach achieves the lowest average epipolar error, reflecting improved geometric accuracy in the estimation of feature correspondences. Additionally, the precision metric shows a marked improvement, with our method consistently outperforming all evaluated baselines. These findings support the hypothesis that contrastive learning for self-supervision from novel-view synthesis for optimizing a feature space adapted from the semantic representations of DINOv2, along with pixel-level refinement, enhances the geometric consistency and reliability of matches in the endoscopic domain.

In contrast to prior deep learning-based models, our pipeline introduces the use of novel-view synthesis during training, serving as a form of supervisory signal. This differentiates our method from approaches such as Farhat~\textit{et al.}~\cite{farhat2023self}, which rely on planar homographies to generate pseudo-ground-truth correspondences, as well as from SuperGlue~\cite{superglue} and LoFTR~\cite{loftr}, which incorporate camera poses and depth maps sampled directly from the training data. Traditional pipelines based on SIFT, combined with modern descriptors such as HyNet or SOSNet, demonstrate competitive performance in terms of inlier ratios; however, their lower precision scores suggest a greater tendency to produce ambiguous or redundant matches.

Moreover, our robustness analysis (Table~\ref{tab:robustness}) indicates that the proposed model generates fewer erroneous correspondences under challenging cross-sequence matching conditions, where no anatomical overlap is expected. This behavior can be attributed to the contrastive optimization of the adaptation module $\Psi$, which is trained to increase the representational dissimilarity between descriptors associated with semantically and spatially distinct image regions. As a result, the similarity scores assigned to incorrect correspondences remain low, effectively reducing the likelihood of false positive matches during the selection process.

From a methodological perspective, the ablation studies in Tables~\ref{tab:component_ablation} and~\ref{tab:training_strategy_ablation} provide insight into the architectural contributions. Notably, the addition of the adaptation layer and resolution upsampling yields progressive improvements, validating each component. The choice of randomly sampling rigid transformations $T_{s \rightarrow t}$ during training further enhances generalization over using dataset camera poses, likely due to increased variation in appearance and viewpoint. This approach enhances the heterogeneity of the supervision signal, thereby improving the generalization capabilities of the trained model. The performance improvements observed validate our decision to incorporate varied transformations in the training process.

Despite the improvements observed, several limitations remain. Primarily, although our approach achieves high precision, it does not yield the highest inlier ratio. This may indicate that the matching density in our method is low, possibly due to its conservative scoring mechanism in rejecting ambiguous matches. While beneficial for robustness, this behavior could limit its applicability in tasks requiring dense correspondence. Another limitation lies in the fact that the supervisory signal is derived by implicitly relying on the accuracy of the depth maps used for novel-view synthesis, which introduces a potential source of geometric noise. Inaccuracies in the estimated depth can propagate to the synthetic views, potentially resulting in misaligned correspondences during training. This constraint highlights the dependency of our self-supervised formulation on the quality of the underlying depth estimation method.

Future work will focus on expanding the scalability and versatility of the proposed framework. Since the training procedure does not rely on annotated camera poses and can be carried out with a single up-to-scale depth map per image pair, the model can, in principle, be trained on arbitrarily long video sequences without any annotation. This flexibility opens the possibility of leveraging diverse surgical endoscopic video datasets where pose annotations are either unavailable or unreliable. Another promising direction involves unifying the currently distinct feature spaces used for semantic representation and geometric matching. Instead of relying on the DINOv2 backbone solely for semantic guidance followed by a dedicated adaptation module for matching, future versions of the model may be trained end-to-end to learn a shared embedding space that simultaneously encodes semantic structure and geometric discriminability, thereby potentially improving feature consistency across tasks.

\section{Conclusions}
This paper presents a self-supervised framework for endoscopic image matching that adapts pretrained visual embeddings for pixel correspondence through contrastive learning. We synthesize a supervisory signal by back-projecting pixels to 3D space and re-projecting them to novel views with randomly generated positions and orientations, creating diverse synthetic ground-truth data for training. This approach preserves the semantic-rich representation of image patches while adapting it to a more matching-oriented domain. Our method shows promising results compared to existing techniques, with reduced epipolar error and improved precision on a standard endoscopic dataset, while demonstrating good robustness in identifying spurious matches. These results suggest potential applications for improving spatial awareness during minimally invasive surgical procedures.

\section*{Acknowledgments}
The authors gratefully acknowledge Asensus Surgical Inc., Durham, NC 27703, USA, for their supervision and support throughout the development of this work. Their technical guidance and domain expertise have been instrumental in shaping the research direction and ensuring the clinical relevance of the proposed methodology.

\bibliographystyle{IEEEtran}
\bibliography{main}

\end{document}